\documentclass{article}

\usepackage{iclr2027_conference}

\usepackage[utf8]{inputenc}     
\usepackage[T1]{fontenc}    	
\usepackage{url}                
\usepackage{booktabs}       	
\usepackage{amsfonts}       	
\usepackage{nicefrac}           
\usepackage{microtype}      	
\usepackage{times}

\usepackage{graphicx}
\usepackage{color}
\usepackage[table]{xcolor}
\usepackage{rotating}
\usepackage{tabularx}
\usepackage{pdflscape}
\usepackage{amsmath,amsthm,amssymb}
\usepackage{algorithm,algorithmic}
\usepackage{bbm,dsfont}
\usepackage{caption}
\usepackage{wrapfig}
\usepackage{cancel}
\usepackage{enumerate,cases}
\usepackage{thmtools,thm-restate}
\usepackage{mathtools}

\usepackage{multirow}
\usepackage{makecell}
\usepackage{setspace}
\usepackage{pifont}
\usepackage{array}
\usepackage{enumitem}
\usepackage{lineno}     
\usepackage{bm}
\usepackage{subcaption}

\allowdisplaybreaks

\usepackage{hyperref}		 
\hypersetup{
}
\usepackage[capitalize,noabbrev]{cleveref}

\crefformat{figure}{Fig.~#2#1#3}       
\crefformat{table}{Tab.~#2#1#3}        
\crefformat{equation}{Eq.~(#2#1#3)}    
\crefformat{section}{Sec.~#2#1#3}      
\crefformat{appendix}{App.~#2#1#3}     
\crefformat{algorithm}{Alg.~#2#1#3}    

\usepackage{silence}
\renewcommand{\phi}{\varphi}
\renewcommand{\epsilon}{\varepsilon}

\newcommand{\el}{\end{flushleft}}
\newcommand{\bl}{\begin{flushleft}}

\newcommand{\squishlisttwo}{
 \begin{list}{$\bullet$}
  { \setlength{\itemsep}{1pt}
     \setlength{\parsep}{0pt}
    \setlength{\topsep}{0pt}
    \setlength{\partopsep}{0pt}
    \setlength{\leftmargin}{1em}
    \setlength{\labelwidth}{1.5em}
    \setlength{\labelsep}{0.5em} } 
}
\newcommand{\squishend}{\end{list}}

\theoremstyle{plain}

\iclrfinalcopy

\makeatletter
\def\@noticestring{}
\makeatother

\title{
Cannot Wait to Verify: Parallel Drafting to Eliminate
Verifier Idle Time in Speculative Decoding
}

\author{
\textbf{Divya Jyoti Bajpai}\textsuperscript{1},
\textbf{Arun Verma}\textsuperscript{2},
\textbf{Manjesh K. Hanawal}\textsuperscript{1}
\\[0.5em]
\textsuperscript{1}Department of IEOR, IIT Bombay, Mumbai, India \\
\textsuperscript{2}Singapore-MIT Alliance for Research and Technology (SMART) Centre, Singapore
\\[0.5em]
\texttt{divyajyoti.bajpai@iitb.ac.in},
\texttt{arun.verma@smart.mit.edu},
\texttt{mhanawal@iitb.ac.in}
}

\begin{document}

\maketitle


\begin{abstract}
    Autoregressive inference in large transformer models is limited by token-level sequential execution, resulting in high latency and low GPU utilization. Speculative decoding (SD) mitigates this bottleneck by introducing a draft-and-verifier model. However, the drafting stage remains sequential, leaving the verifier idle and underutilizing compute. To address this, we propose a plug-and-play drafting adaptive drafting method that parallelizes inference across transformer layers during draft generation. Starting from a shared hidden state, our approach evaluates multiple layers in parallel and verifies consistency by measuring divergence in hidden-state space. Layer-wise draft states are accepted when deviations are small; otherwise, a correction step is applied. By exploiting redundancy across transformer layers, our method accelerates draft generation and improves hardware utilization without modifying architectures or requiring retraining. Experimental results on NLP tasks show that our method speeds up the drafting stage by up to $1.65\times$ with minimal degradation in accuracy, while maintaining the same performance as SD. The anonymized source code for our method can be found at \url{https://anonymous.4open.science/r/PDEVIS-3EDD/README.md}.
\end{abstract}

\section{Introduction}
\label{sec:introduction}


Large Language Models (LLMs)~\citep{minaee2024large}  
have achieved impressive performance across various tasks 
(reasoning, generation, decision-making, etc.) largely driven by aggressive scaling in depth and parameter count. This scaling has made autoregressive decoding a primary bottleneck in deployment, as each generated token requires a full forward pass through all transformer layers, enforcing strict token-wise sequential execution. This results in high inference latency, limiting the practicality of large models in latency-sensitive settings \citep{bajpai2025beyond, cai2024medusa}.


Numerous approaches have been proposed to mitigate inference latency, including pruning \citep{vadera2022methods}, quantization \citep{gholami2022survey}, distillation \citep{xu2024survey}, and early-exit \citep{rahmath2024early} mechanisms. While effective in reducing computation, these methods typically trade accuracy for efficiency by altering model architectures, compressing representations, or terminating computation prematurely. In contrast, decoding-time methods such as speculative decoding (SD) \citep{hu2025speculative} achieve acceleration without sacrificing output quality by reducing sequential dependence: a lightweight draft model rapidly generates candidate tokens autoregressively, which are then verified in parallel by a larger, more accurate model.

\begin{figure}[t]
    \centering
    \begin{subfigure}{0.47\linewidth}
        \centering
        \includegraphics[width=\linewidth]{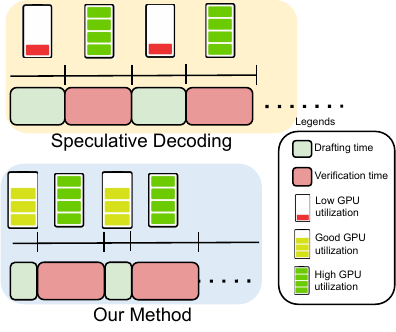}
        \caption{GPU utilization in SD vs our method. SD shows verifier idle time during drafting; our method reduces idle time and improves parallel efficiency.}
        \label{fig:idle_time}
    \end{subfigure}
    \hfill
    \begin{subfigure}{0.47\linewidth}
        \centering
        \includegraphics[width=\linewidth]{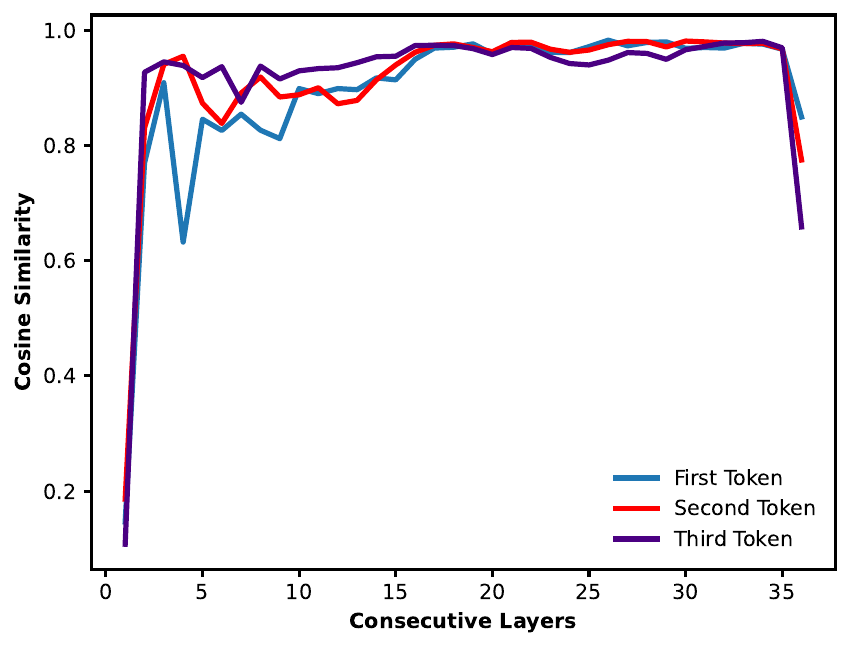}
        \caption{Layer-wise similarity between consecutive transformer layers in the LLaMA-3B draft model across multiple decoding timesteps.}
        \label{fig:layer_similarity}
    \end{subfigure}
    \vspace{-4mm}
\end{figure}
Despite its success, SD suffers from two fundamental inefficiencies that have received limited attention. First, the drafting stage remains fully autoregressive, forcing the verifier to idle until a batch of draft tokens is produced. This leads to substantial underutilization of compute hardware and limits the achievable speedup (see Figure \ref{fig:idle_time}). Second, SD is governed by a trade-off in the drafting stage: a slightly bigger/slower, more accurate draft model improves token acceptance and reduces verification frequency, but incurs higher autoregressive latency during drafting, while a smaller/faster, less accurate drafter lowers per-step drafting cost at the expense of increased rejections and reduced parallel efficiency. Together, these effects make the sequential nature of drafting the dominant bottleneck in SD, impacting the overall system efficiency.

Prior work that attempts to improve the drafting process relies on weaker or partially executed draft models \citep{wu2025easyspec}, assuming that verification can cheaply correct drafting errors. However, frequent rejections limit end-to-end speedup in such methods. Moreover, existing approaches employ static drafting strategies \citep{leviathan2023fast, elhoushi2024layer}, allocating a fixed speedup method regardless of input difficulty. These limitations motivate an adaptive drafting mechanism that dynamically balances speed and accuracy, accelerating draft generation for easy inputs while selectively increasing computation for harder cases to maximize acceptance.

\begin{figure}[!ht]
    \vspace{-1mm}
    \centering
    \includegraphics[scale=0.59]{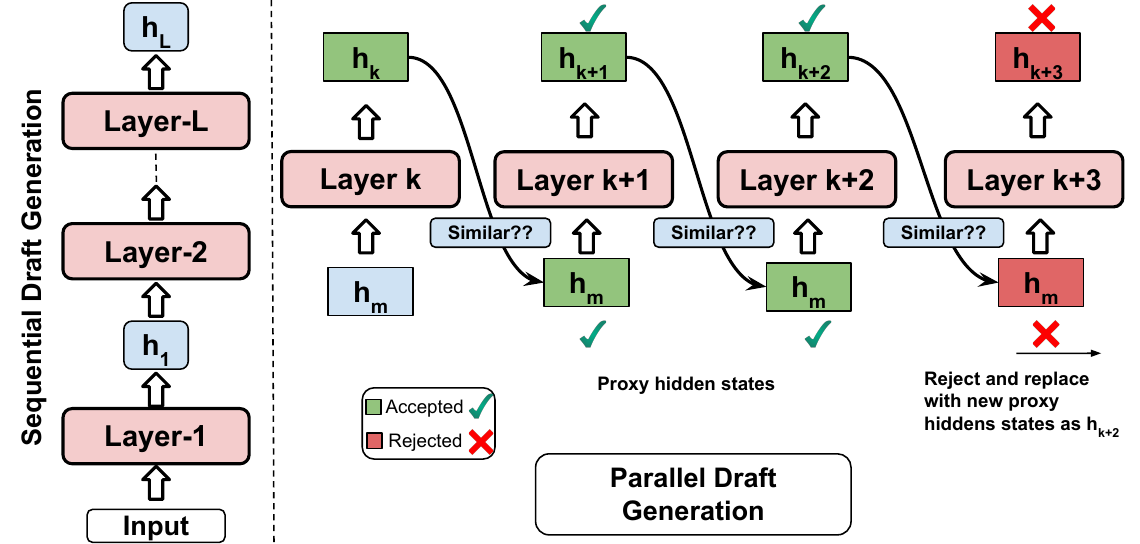}
    \caption{\textbf{Vanilla versus Adaptive Layer-parallel Drafting (ALD).} Vanilla drafting executes layers sequentially, leaving parallel compute underutilized. ALD evaluates future layer transformations concurrently using proxy hidden states and adaptively controls their advancement through a similarity check, reducing drafting latency while preserving draft quality.}
    \label{fig:main}
    \vspace{-2mm}
\end{figure}

To overcome this limitation, we introduce a \textit{layer-parallel drafting mechanism} that adaptively and efficiently accelerates the draft-generation stage. Instead of executing transformer layers of the draft model strictly sequentially, our method evaluates multiple layers concurrently by propagating from the most recently available hidden state. A lightweight similarity check continuously monitors divergence in hidden-state space and triggers recomputation only when the proposed hidden state deviates substantially from the true hidden state. When divergence remains small, intermediate layers are effectively skipped by reusing hidden representations, allowing the draft model to advance rapidly with minimal additional computation and approximation error. We detail our method in Figure \ref{fig:main}.

Our approach is motivated by empirical evidence that transformer models exhibit substantial \textit{layer-wise redundancy} during inference. As seen in Fig.~\ref{fig:layer_similarity}, many inputs do not require the full depth of the network to produce stable representations. By exploiting this redundancy at the layer level, we reduce the \textit{effective sequential depth} of the draft model while retaining the ability to self-correct when necessary. Crucially, the proposed method is \textit{plug-and-play}, requires no architectural modifications or retraining, and integrates seamlessly with existing speculative decoding pipelines. As a result, it unlocks additional parallelism during inference, improves hardware utilization, and enables latency reductions beyond what standard speculative decoding can achieve, without compromising quality.

Empirical results demonstrate that layer-parallel drafting achieves substantial speedups while fully preserving draft quality, yielding comparable token acceptance rates during verification by the larger model. These features make our method a practical enhancement for real-time, large-scale autoregressive inference. In summary, the key contributions of this paper are as follows:
\vspace{-2mm}
\begin{itemize}
    \item We introduce a plug-and-play method that parallelizes transformer layers during speculative drafting, reducing sequential bottlenecks and draft latency.
    \item We exploit inter-layer redundancy to adaptively reduce sequential depth while preserving draft quality, without retraining or architecture changes.
    \item Our method improves the efficiency of speculative decoding while maintaining high token acceptance and output quality, thereby enabling efficient autoregressive inference.
    \item Experimentally, across diverse tasks, we achieve up to $>1.65\times$ drafter speedup with minimal loss in draft accuracy while preserving SD's end-to-end performance.
\end{itemize}

\section{Related Work}
\label{sec:related_work}

Speculative Decoding (SD) \citep{stern2018blockwise, sun2021instantaneous, leviathan2023fast, chen2023accelerating, li2024eagle, li2024eagle1, li2025eagle} has been one of the major advances in faster inference methods for autoregressive models without performance loss. Several prior works have studied ways to improve SD. Given the pivotal role of distributional consistency between the draft and target models, researchers have focused on aligning the draft model with the target model through additional training \citep{cai2024medusa, ankner2024hydra}. Methods that focus on the drafting stage are mainly training the drafters or using self-speculation \citep{zhang2024learning, elhoushi2024layer}. Specinfer \citep{miao2024specinfer} uses boost-trained draft models and tree-structured attention for efficient speculation. 

Medusa \citep{cai2024medusa} trains a set of extra MLP heads for future token prediction using the original LLM's features. GLIDE and CAPE \citep{du2024glide} reuse the target model's KV-cache, while methods \citep{hooper2025speed, yang2023predictive, monea2023pass, li2024nearest, yi2024generation, liu2024kangaroo, sun2024triforce, svirschevski2024specexec} such as Self-Speculative Decoding \citep{zhang2023draft, elhoushi2024layer} combine early exits or layer skipping with speculative decoding. LayerSkip \citep{elhoushi2024layer} and Draft \& Verify \citep{zhang2023draft} use some initial layers of the same model as a draft. Lookahead \citep{fu2024break} uses n-gram Jacobi decoding to increase acceptance rates. REST \citep{he2024rest} leverages a data storage for retrieval speculation. 

Recent works also explore optimized tree-structured attention mechanisms. SpecExec \citep{svirschevski2024specexec} takes the most probable continuations from the draft model to build a cache tree, improving inference efficiency on resource-constrained GPUs. OPT-Tree \citep{wang2025opt} proposes an adaptive and scalable draft-tree construction algorithm that maximizes the expected acceptance length. Beyond the drafter itself, other studies investigate stage-level parallelism and theoretical explanations of SD. For instance, PEARL \citep{liu2024pearl} overlaps the drafting and verification stages to achieve stage-level pipelines, while SpecTr \citep{sun2023spectr} adopts optimal transport theory to provide a theoretical explanation and optimization. EasySpec \citep{wu2025easyspec} produces fuzzy speculative drafts in parallel, hypothesizing that drafting was never meant to be precise, thereby leading to lower acceptance rates and limiting efficiency.

In contrast, we introduce a method that challenges the hypothesis that faster draft generation necessarily leads to lower acceptance rates due to reduced draft quality. We achieve this not only by executing draft layers in parallel but also by performing explicit quality checks to ensure that the drafter's overall generation quality is not compromised and that the token acceptance rate remains stable. 
To the best of our knowledge, we are the first to propose an adaptive drafting mechanism that improves draft model quality without affecting the token acceptance rate.


\section{Setup}
\label{sec:problem}

In this section, we introduce Speculative Decoding (SD) and the distributed inference setup.

\subsection{Speculative Decoding}\label{sec:speculative_decoding}
SD \citep{leviathan2023fast} decomposes decoding into two alternating components: a fast \emph{drafting} step executed by a small draft model and a \emph{verification} step carried out by the target model. This design enables the verification of multiple candidate tokens generated by the draft model in a single iteration.
Let $x_{1:t}$ denote the generated prefix at decoding step $t$. In the drafting step, a lightweight model $M_q$ autoregressively generates a sequence of $k$ speculative tokens $ x'_{t+1:t+k} = (x'_{t+1}, \ldots, x'_{t+k}),$
together with draft output distributions $\{q_i\}_{i=t+1}^{t+k}$, where
$q_i(\cdot) = M_q(\cdot|x_{1:t}, x'_{t+1:t+i-1})$.
The draft model is chosen to be computationally efficient, allowing speculative exploration of future tokens at low cost.

The candidate tokens are then verified by the target model $M_p$ in a single forward pass. Conditioning on the concatenated sequence $(x_{1:t}, x'_{t+1:t+k})$, the verifier produces distributions $\{p_i\}_{i=t+1}^{t+k+1},$, where
$p_i(\cdot) = M_p(\cdot |x_{1:t}, x'_{t+1:t+i-1}),$
including one additional distribution beyond the drafting horizon.

Each speculative token $x'_i$ is accepted independently according to an importance-sampling-based acceptance rule. Specifically, the acceptance probability is defined as
$$\alpha_i = \min\!\left(1, \frac{p_i(x'_i)}{q_i(x'_i)}\right).$$
Let $m \leq k$ denote the largest index such that tokens $x'_{t+1}, \ldots, x'_{t+m}$ are consecutively accepted. These tokens are appended to the output sequence, yielding an updated prefix $x_{1:t+m}$.

To correct the $(m+1)^{th}$ rejected token, decoding proceeds by sampling an additional token from a distribution derived from the verifier outputs. If $m < k$, this distribution is constructed by subtracting the draft model's probability mass from the verifier distribution and re-normalizing, otherwise the verifier distribution at the subsequent position is used:
\begin{equation}
p =
\left\{
\begin{array}{ll}
\mathrm{norm}\!\left(\max\!\left(0,\, p_{t+m+1} - q_{t+m+1}\right)\right), & m < k, \\[4pt]
p_{t+m+1}, & m = k .
\end{array}
\right.
\end{equation}

By iteratively repeating the drafting and verification steps, speculative decoding reduces the number of expensive autoregressive calls required during inference. The acceptance-rejection mechanism guarantees that the resulting sequence is distributed exactly according to the target model, thereby preserving correctness \citep{leviathan2023fast}.

However, this setup assumes a sufficient amount of compute that can perform a parallel verification from a verifier model. This compute remains idle during the draft generation process, waiting for the draft tokens available. This makes the overall compute underutilized in the SD method.

\subsection{Layer Skipping}
Layer skipping is a widely studied strategy for reducing inference latency in LLMs, most commonly through early-exit mechanisms \citep{bajpai2025survey}. The core premise is that not all inputs require full-depth computation: when sufficiently informative representations are formed at intermediate layers, further processing may be redundant. By adaptively terminating inference, such methods can substantially reduce average computational cost.

\textbf{Why Naive Layer Skipping Fails in SD.}
Although skipping layers can reduce draft computation, naive layer-skipping strategies are ill-suited for accelerating the drafting stage in SD. Decisions based on heuristic confidence signals can be unreliable, particularly for ambiguous or distribution-shifted inputs. Such inaccuracies directly reduce the token acceptance rate during verification, causing speculative tokens to be rejected more frequently and increasing the number of draft-verification iterations, affecting overall speedup.

\textbf{Need for Acceptance-Aware Skipping.}
Effective skipping strategies in SD must explicitly account for their downstream impact on verifier acceptance. Skipping decisions should preserve draft model performance, ensuring that speculative tokens remain compatible with the verifier's evaluation. Without such acceptance-aware constraints, aggressive skipping degrades overall system efficiency and speedup despite lowering the per-step computational cost of the draft model.

\textbf{Motivation for Reliable, Resource-Aware Skipping.}
Rather than terminating computation early or relying on heuristic confidence thresholds, the draft model should selectively eliminate redundant computation while explicitly constraining representational deviation. These considerations motivate a principled mechanism that exploits otherwise idle computational resources during the drafting stage. We next discuss a mechanism that accelerates candidate token generations while maintaining high acceptance rates and stable SD dynamics.

\section{Methodology} In this section, we detail our method, where we perform a layer parallel strategy during the drafting process and then verify the generated draft tokens.

\subsection{Drafting Stage}
\label{sec:drafting}

Let the draft model $M_q$ consist of $L$ transformer layers $\{f_0,\ldots,f_{L-1}\}$, and let $h_0$ denote the hidden representation produced by the embedding layer. Under standard execution, the hidden states are computed sequentially as
$$
    h_{k+1}^{*}=f_k(h_k^{*}), \qquad k=0,\ldots,L-1,
$$
where $h_k^{*}$ denotes the hidden state produced by the exact sequential execution of the draft model.

Our objective is to reduce the sequential depth of this computation by evaluating multiple future layer transformations concurrently. ALD maintains a committed hidden state $\bar{h}_m$ at depth $m$, where $m$ denotes the deepest layer reached by the current committed execution. Initially, $m=0$ and $\bar{h}_0=h_0$.

For a parallel window beginning at depth $m$, we define its endpoint as
$$
    K_m=\min(m+K,L),
$$
where $K$ is the maximum parallel window size. ALD constructs a shared proxy input for the layers in this window by reusing the current committed representation:
$$
    \tilde{h}_k=\bar{h}_m, \qquad k=m,\ldots,K_m-1.
$$
The corresponding candidate states are evaluated independently and in parallel:
$$
    u_{k+1}=f_k(\tilde{h}_k)=f_k(\bar{h}_m), \qquad k=m,\ldots,K_m-1.
$$
The first candidate $u_{m+1}=f_m(\bar{h}_m)$ is an exact one-layer advancement from the current committed state. Candidates farther into the window replace their usual sequential inputs with the shared proxy $\bar{h}_m$, thereby exposing layer-level parallelism.

ALD determines how far the parallel window can be advanced using hidden-state similarity. For each subsequent candidate transition $k\in\{m+1,\ldots,K_m-1\}$, we compute
$$
    s_k=\frac{\left\langle u_{k+1},\bar{h}_m\right\rangle}{\left\|u_{k+1}\right\|_2\left\|\bar{h}_m\right\|_2}.
$$
The cosine similarity serves as an online signal for the stability of the proxy execution relative to the current committed representation. A high value indicates that the corresponding layer transformation largely preserves the direction of the committed representation, whereas a lower value indicates a larger representational change. The threshold $\tau$ therefore controls the trade-off between parallel advancement and correction frequency: a higher threshold makes ALD more conservative, whereas a lower threshold permits more aggressive parallel execution.

Let the first layer whose proxy candidate fails the similarity criterion be
$$
    j=\min\{k\in\{m+1,\ldots,K_m-1\}:s_k<\tau\},
$$
when such a layer exists. If the criterion first fails at layer $j$, the candidate $u_{j+1}$ is rejected, and $u_j$ is retained as the last accepted candidate state. ALD then updates
$$
    m\leftarrow j,\qquad \bar{h}_m\leftarrow u_j,
$$
and constructs a new parallel window from this committed representation. If no rejection occurs in the current window, the window is accepted through its endpoint and
$$
    m\leftarrow K_m,\qquad \bar{h}_m\leftarrow u_{K_m}.
$$

The procedure is repeated until $m=L$, after which the final committed state $\bar{h}_L$ is used to obtain the draft-model output distribution for the current decoding step. By evaluating multiple future layer transformations concurrently and adaptively advancing the committed state, ALD reduces the effective sequential depth of draft-model execution while preserving the original transformer architecture and parameters. The method, therefore, trades additional parallel computation for reduced sequential drafting latency, with the similarity criterion determining how aggressively the proxy execution is advanced through the remaining layers.

\begin{algorithm}[t]
\caption{Adaptive Layer-parallel Drafting (ALD)}
\label{alg:ald}
\begin{algorithmic}[1]
\REQUIRE Draft model layers $\{f_k\}_{k=0}^{L-1}$, initial hidden state $h_0$, similarity threshold $\tau$, maximum parallel window $K$
\ENSURE Final draft hidden state $\bar{h}_L$

\STATE Initialize committed depth $m\leftarrow0$ and committed state $\bar{h}_m\leftarrow h_0$

\WHILE{$m<L$}
\STATE $K_m\leftarrow\min(m+K,L)$

\STATE \textbf{Proxy construction:}
\FOR{$k=m,\ldots,K_m-1$}
\STATE $\tilde{h}_k\leftarrow\bar{h}_m$
\ENDFOR

\STATE \textbf{Parallel candidate evaluation:}
\FOR{$k=m,\ldots,K_m-1$ \textbf{ in parallel}}
\STATE $u_{k+1}\leftarrow f_k(\tilde{h}_k)$
\ENDFOR

\STATE \textbf{Similarity check:}
\FOR{$k=m+1,\ldots,K_m-1$}
\STATE $s_k\leftarrow\dfrac{\langle u_{k+1},\bar{h}_m\rangle}{\|u_{k+1}\|_2\|\bar{h}_m\|_2}$
\ENDFOR

\STATE $j\leftarrow\min\{k\in\{m+1,\ldots,K_m-1\}:s_k<\tau\}$, if such an index exists

\IF{$j$ exists}
\STATE $m\leftarrow j$; $\bar{h}_m\leftarrow u_j$
\ELSE
\STATE $m\leftarrow K_m$; $\bar{h}_m\leftarrow u_{K_m}$
\ENDIF
\ENDWHILE

\RETURN $\bar{h}_L$
\end{algorithmic}
\end{algorithm}

We refer to the resulting procedure as \emph{Adaptive Layer-parallel Drafting (ALD)}. Because ALD can advance across multiple layers using proxy evaluations, a committed ALD representation may deviate from the representation obtained by sequentially composing the corresponding transformer layers. We next characterize how these local approximation errors propagate to the final hidden representation.

\begin{restatable}{thm}{speculativeBound}
\label{thm:speculative_bound}
Let $0=m_0<\cdots<m_T=L$ be the depths committed by ALD, and define $F_r:=f_{m_{r+1}-1}\circ\cdots\circ f_{m_r}$ with Lipschitz constant $\Lambda_r:=\prod_{k=m_r}^{m_{r+1}-1}L_k$. Let $H_0^*=h_0$, $H_{r+1}^*=F_r(H_r^*)$, and define the local ALD approximation error $\epsilon_r:=\|\bar{h}_{m_{r+1}}-F_r(\bar{h}_{m_r})\|_2$. Then
\begin{equation}
\|\bar{h}_L-H_T^*\|_2\leq\sum_{r=0}^{T-1}\left(\prod_{q=r+1}^{T-1}\Lambda_q\right)\epsilon_r,
\end{equation}
where an empty product equals $1$.
\end{restatable}

Theorem~\ref{thm:speculative_bound} shows that the final deviation depends on the local approximation errors introduced when ALD advances between committed depths and on their propagation through subsequent layer blocks. Smaller proxy deviations contribute less to the final error, whereas errors introduced earlier in the execution can be propagated through more subsequent transformations.

The factors $\Lambda_q$ characterize the worst-case amplification of a local error by the exact layer compositions between successive committed depths. Although their products can grow in the worst case, their practical magnitude depends on the local stability of the transformer mappings along the inference trajectory. Residual connections and normalization can mitigate perturbation growth over moderate layer horizons; we therefore interpret the bound as a worst-case characterization of error propagation rather than assuming uniformly small Lipschitz constants.

Empirically, the draft models considered in this work exhibit substantial layer-wise representational similarity, motivating cosine similarity as the online signal used by ALD to identify stable regions for parallel proxy execution. Theorem~\ref{thm:speculative_bound} complements this observation by characterizing how resulting local approximation errors propagate through subsequent layer blocks. In Section~\ref{sec:experiments}, we evaluate their practical effect through draft quality, verifier acceptance, and end-to-end decoding speedup.

\subsection{Verification Process in Target Model}
Once the speculative token sequence is generated by the draft model, it is passed to the target model for verification. We use the standard speculative decoding verification procedure described in Section~\ref{sec:speculative_decoding}. The target model evaluates the speculative sequence in a single forward pass and produces the target distributions for all candidate positions. The speculative tokens are then processed in order using the standard acceptance-rejection rule until the first rejection occurs. If a speculative token is rejected, the next token is sampled from the corrected distribution defined in Section~\ref{sec:speculative_decoding}; if all speculative tokens are accepted, an additional token is sampled from the target model distribution at the subsequent position. Consequently, ALD modifies only the draft-generation procedure, while target-model verification and the standard SD correction mechanism remain unchanged.

\subsection{KV-Cache Alignment}
\label{sec:kv_alignment}

\textbf{KV-cache error accumulation.} During layer-parallel drafting, ALD generates speculative tokens, draft distributions, and the associated key-value (KV) states using proxy hidden representations rather than fully sequential layer propagation. While errors in speculative tokens and their output distributions are resolved at the end of each draft--verify iteration through standard speculative verification, approximate KV states can persist across decoding iterations if they are retained in the cache. Reusing such approximate states as conditioning context may cause representation errors to accumulate over decoding steps, potentially degrading draft quality and verifier acceptance.

\textbf{Bonus calibration for KV-cache recovery.} To prevent persistent KV-cache misalignment, we adopt the bonus calibration strategy of \citet{wu2025easyspec}. After each draft--verify iteration, the approximate KV entries associated with the speculative execution are discarded. The accepted speculative tokens, together with the bonus token produced by the target model, are then reprocessed by the draft model to reconstruct the KV states required for the next drafting iteration. This restores the draft-model cache before subsequent speculative generation, preventing approximation errors in the KV cache from accumulating across iterations. The calibration introduces an additional draft-model pass, but it can process the accepted segment in parallel and is included in our end-to-end runtime measurements.

\section{Experiments}
\label{sec:experiments}

In this section, we present the experimental setup and a detailed analysis of all empirical results.

\begin{table*}[t]
\centering
\caption{Acceptance rate, end-to-end (E2E) speedup, and draft time across datasets. E2E speedup is relative to autoregressive target-model decoding; draft time is in seconds per 100 tokens, with parenthetical values denoting speedup over standard SD.}
\label{tab:main-results}
\vspace{-2.25mm}
\small
\setlength{\tabcolsep}{5pt}
\begin{tabular}{llcccccccc}
\toprule
\multirow{2}{*}{Dataset} & \multirow{2}{*}{Method} & \multicolumn{3}{c}{Qwen-4B $\rightarrow$ 30B} & \multicolumn{3}{c}{LLaMA-3B $\rightarrow$ 70B} \\
\cmidrule(lr){3-5} \cmidrule(lr){6-8}
& & Accept. & E2E Spd. & Draft Time & Accept. & E2E Spd. & Draft Time \\
& & & & (s/100 tok.) & & & (s/100 tok.) \\
\midrule
\multirow{4}{*}{WMT}
& SD            & 0.40 & 2.47$\times$ & 30.2 (1.00$\times$) & 0.50 & 1.91$\times$ & 48.5 (1.00$\times$) \\
& EasySpec      & 0.32 & 2.72$\times$ & 19.4 (1.56$\times$) & 0.44 & 2.08$\times$ & 34.4 (1.41$\times$) \\
& LayerSkip     & 0.25 & 2.54$\times$ & 18.1 (1.67$\times$) & 0.31 & 1.96$\times$ & 30.8 (1.57$\times$) \\
& \textbf{ALD} & \textbf{0.38} & \textbf{2.95$\times$} & 18.7 (1.62$\times$) & \textbf{0.46} & \textbf{2.29$\times$} & 32.5 (1.49$\times$) \\
\midrule
\multirow{4}{*}{MMLU}
& SD            & 0.39 & 2.05$\times$ & 31.2 (1.00$\times$) & 0.55 & 1.96$\times$ & 48.9 (1.00$\times$) \\
& EasySpec      & 0.31 & 2.49$\times$ & 18.8 (1.66$\times$) & 0.42 & 2.24$\times$ & 33.9 (1.44$\times$) \\
& LayerSkip     & 0.23 & 2.36$\times$ & 17.5 (1.78$\times$) & 0.34 & 2.09$\times$ & 31.2 (1.57$\times$) \\
& \textbf{ALD} & \textbf{0.36} & \textbf{2.77$\times$} & 17.2 (1.81$\times$) & \textbf{0.51} & \textbf{2.65$\times$} & 30.7 (1.59$\times$) \\
\midrule
\multirow{4}{*}{MGSM}
& SD            & 0.35 & 1.85$\times$ & 30.8 (1.00$\times$) & 0.61 & 2.32$\times$ & 47.8 (1.00$\times$) \\
& EasySpec      & 0.29 & 2.30$\times$ & 19.2 (1.60$\times$) & 0.56 & 2.61$\times$ & 34.5 (1.39$\times$) \\
& LayerSkip     & 0.20 & 2.05$\times$ & 18.4 (1.67$\times$) & 0.41 & 2.50$\times$ & 31.8 (1.50$\times$) \\
& \textbf{ALD} & \textbf{0.35} & \textbf{2.61$\times$} & 19.5 (1.58$\times$) & \textbf{0.60} & \textbf{2.79$\times$} & 34.1 (1.40$\times$) \\
\midrule
\multirow{4}{*}{IF-Eval}
& SD            & 0.36 & 1.92$\times$ & 31.4 (1.00$\times$) & 0.39 & 1.71$\times$ & 48.2 (1.00$\times$) \\
& EasySpec      & 0.29 & 2.52$\times$ & 19.6 (1.60$\times$) & 0.32 & 2.04$\times$ & 34.8 (1.39$\times$) \\
& LayerSkip     & 0.18 & 2.38$\times$ & 17.8 (1.76$\times$) & 0.28 & 1.97$\times$ & 30.4 (1.59$\times$) \\
& \textbf{ALD} & \textbf{0.35} & \textbf{2.87$\times$} & 19.2 (1.64$\times$) & \textbf{0.35} & \textbf{2.21$\times$} & 30.8 (1.57$\times$) \\
\midrule
\multirow{4}{*}{Math}
& SD            & 0.42 & 2.19$\times$ & 29.9 (1.00$\times$) & 0.53 & 2.06$\times$ & 48.5 (1.00$\times$) \\
& EasySpec      & 0.35 & 3.25$\times$ & 18.9 (1.58$\times$) & 0.48 & 2.45$\times$ & 34.0 (1.43$\times$) \\
& LayerSkip     & 0.28 & 2.76$\times$ & 18.0 (1.66$\times$) & 0.32 & 2.38$\times$ & 31.9 (1.52$\times$) \\
& \textbf{ALD} & \textbf{0.39} & \textbf{3.75$\times$} & 17.5 (1.71$\times$) & \textbf{0.52} & \textbf{2.59$\times$} & 33.5 (1.45$\times$) \\
\bottomrule
\end{tabular}
\vspace{-5.5mm}
\end{table*}

\textbf{Datasets.} We evaluate our method on a diverse set of benchmarks spanning multiple NLP tasks. For machine translation, we use WMT~\citep{kocmi2024findings}. General language understanding and reasoning are evaluated on MMLU~\citep{hendrycks2009measuring}. Mathematical reasoning is assessed using MGSM~\citep{shi2022language} and the Math dataset, covering multilingual grade-school problems and more challenging multi-step mathematical reasoning, respectively. Instruction following ability is evaluated using IF-Eval~\citep{zhou2023instruction}. Finally, we use SpecBench~\citep{xia2024unlocking} to evaluate speculative decoding across multiple generation tasks. Together, these benchmarks cover translation, language understanding, mathematical reasoning, instruction following, and SD efficiency.

\textbf{Metrics.} We evaluate efficiency using end-to-end decoding speedup relative to standard autoregressive target-model decoding, draft-generation latency $d$ (seconds per 100 draft tokens), and speculative token acceptance rate. Speedup captures the overall decoding acceleration, while $d$ isolates the drafting cost. The acceptance rate measures draft--target agreement and directly affects speculative decoding efficiency. Since ALD retains the standard acceptance--rejection mechanism, the final output distribution remains governed by the target model~\citep{leviathan2023fast}.

\textbf{Configurations.}\label{sec:hyperparameters} All experiments are conducted on a system equipped with four NVIDIA RTX A6000 GPUs. To enable layer-parallel draft execution, we use batched inference to process proxy layer evaluations concurrently. The speculative draft length, defined as the number of tokens generated between consecutive verification steps, is fixed to $6$ across all experiments. We use a similarity threshold of $\tau=0.95$ for all models and datasets, motivated by the high layer-wise representational similarity observed during draft-model inference (Figure~\ref{fig:layer_similarity}). We use the same threshold across all experiments rather than tuning it separately for individual datasets. The effect of varying $\tau$ on the speed--quality trade-off is analyzed in Appendix~\ref{sec:speedvsaccuracy} and Figure~\ref{fig:speed_vs_quality}.

For the maximum parallel window size, we set $K=10$ across all experiments. Here, $K$ determines the maximum number of future layer transformations evaluated concurrently from the current committed representation. We choose $K=10$ based on the observed range over which ALD can effectively exploit layer-level parallelism before the similarity criterion typically triggers an earlier update of the committed state. Increasing $K$ exposes additional parallelism but yields diminishing returns because candidates farther into the window are more likely to violate the similarity criterion. In our experiments, increasing $K$ beyond $10$, including using the full remaining depth, provides only limited additional speedup.

\textbf{Baselines.} We compare ALD against the following representative speculative decoding approaches.

\textbf{1) Speculative Decoding (SD).} Standard SD using an autoregressive draft model to generate candidate tokens, followed by parallel verification with the target model~\citep{leviathan2023fast}.

\textbf{2) EasySpec.} A training-free layer-parallel speculative decoding method that relaxes sequential dependencies within the draft model to generate approximate draft representations in parallel, followed by KV-cache calibration~\citep{wu2025easyspec}. Unlike ALD, EasySpec uses a fixed layer-parallel execution strategy rather than adapting the parallel advancement using online hidden-state similarity.

\textbf{3) LayerSkip.} A self-speculative decoding approach that combines layer skipping and early-exit prediction, using a reduced subset of the target model's layers for drafting and the full model for verification~\citep{elhoushi2024layer}.

\textbf{4) ALD (Ours).} Our training-free adaptive layer-parallel drafting method evaluates multiple future layer transformations concurrently from the current committed representation and dynamically controls the advancement of the proxy execution using hidden-state similarity.

For all baselines, we use the hyperparameters reported in their official implementations unless otherwise stated. ALD uses the configuration described in Section~\ref{sec:hyperparameters}.

\subsection{Results}
Table~\ref{tab:main-results} and Table~\ref{tab:task-level-results} compare ALD with the baseline methods across model pairs and tasks. ALD consistently improves end-to-end decoding speedup over standard SD and the compared draft-acceleration baselines, while retaining acceptance rates close to those of standard SD. These results highlight the central trade-off in speculative decoding: reducing drafting latency alone is insufficient if the resulting approximation substantially lowers verifier acceptance.

\textbf{Comparison with Existing Baselines.} EasySpec reduces drafting latency through approximate layer-parallel execution, but its lower acceptance rates can offset part of the resulting draft-time savings. LayerSkip often achieves low drafting latency through more aggressive computation reduction, but this is accompanied by a larger decrease in acceptance. In contrast, ALD uses the online similarity criterion to control the extent of proxy execution, providing substantial draft acceleration while preserving acceptance rates closer to standard SD. For example, on MMLU with Qwen-4B $\rightarrow$ Qwen-30B, ALD achieves a $2.77\times$ end-to-end speedup with an acceptance rate of $0.36$, compared with $2.49\times$ and $0.31$ for EasySpec, and $2.36\times$ and $0.23$ for LayerSkip.

\textbf{Benefits of Adaptive Layer-Parallel Drafting.} The results demonstrate the benefit of adapting the amount of layer-parallel execution rather than applying a fixed approximation uniformly. When the hidden-state similarity remains high, ALD can advance farther through the draft model using parallel proxy evaluations; when the similarity criterion is violated, the committed state is updated earlier. This input-dependent behavior allows ALD to balance drafting latency against the quality of the speculative proposals without requiring retraining or architectural modification.

\textbf{End-to-End Efficiency and Draft Runtime.} ALD does not always have the lowest standalone drafting time, as some baselines use cheaper draft computation. However, its higher acceptance can compensate at the system level. For example, on WMT with Qwen-4B $\rightarrow$ Qwen-30B, LayerSkip has a slightly lower draft time than ALD ($18.1$ vs.\ $18.7$ seconds per 100 draft tokens), but ALD achieves higher end-to-end speedup ($2.95\times$ vs.\ $2.54\times$) and acceptance ($0.38$ vs.\ $0.25$). This highlights that end-to-end SD efficiency depends jointly on draft cost and verifier acceptance.

Overall, ALD provides a favorable balance between draft acceleration and verifier acceptance across the evaluated model pairs and tasks. The results indicate that adaptively controlling layer-parallel proxy execution can improve end-to-end speculative decoding efficiency while limiting the acceptance degradation associated with more aggressive draft approximations.



\section{Conclusion}
\label{sec:conclusion}

In this work, we propose a draft generation method that adaptively speeds up the draft model using the layer-parallel drafting mechanism, which reduces the effective sequential depth of the model without retraining, architectural modifications, or loss of draft fidelity. 
Our method adaptively exploits this redundancy by skipping computation for easy inputs while retaining the ability to revert to full execution for harder cases, thereby preserving high verifier acceptance rates and stable decoding dynamics. As a plug-and-play enhancement to speculative decoding, the proposed approach unlocks additional parallelism, significantly reduces verifier idle time, and improves overall throughput.



\clearpage

\subsection*{AI Use Statement}

In this work, we used generative AI tools solely to assist with writing and language refinement, specifically, improving the clarity, grammar, phrasing, and readability of author-drafted text. All technical content, including the problem formulation, methodology, theoretical results and proofs, experimental design, and analysis, was conceived, developed, and written by the authors. AI tools were not used to generate research ideas, derive or verify proofs, produce experimental results, write or debug code, conduct the literature review, or create any figures, tables, or data. Any such tasks requiring disclosure under the policy are therefore not applicable to this work.

We have reviewed all AI-assisted text. Every passage edited or rephrased with AI assistance was checked by the authors for factual accuracy, correct representation of our claims, and consistency with the underlying technical content, and was revised as needed. We take full responsibility for the final content of this work, including all text, claims, and artifacts.

\subsection*{Ethics Statement}

This work studies inference-time acceleration for large language models and does not involve human subjects, collection of personal or sensitive information, or the introduction of new datasets containing such information. Our experiments use existing publicly available benchmarks and pretrained language models. The proposed method is a general-purpose inference optimization and does not modify the underlying capabilities, training data, or safety mechanisms of the target language model. As with other techniques that reduce the computational cost of large-model inference, improved efficiency may facilitate broader deployment of language models and therefore inherits the potential societal benefits and risks associated with the underlying models and their applications. We encourage deployment in accordance with the licenses, usage policies, safety requirements, and applicable legal requirements of the underlying models and datasets.

\subsection*{Reproducibility Statement}
We provide the information required to reproduce the proposed method and evaluate its main claims throughout the paper and supplementary material. Section~\ref{sec:drafting} formally specifies Adaptive Layer-parallel Drafting (ALD) and provides its algorithm, including proxy construction, parallel layer evaluation, the similarity criterion, and adaptive advancement. Section~\ref{sec:hyperparameters} reports the principal experimental configuration and hyperparameters. Additional implementation and evaluation details are provided in Appendix~\ref{sec:implementation_details}, including the speculative draft length, similarity threshold, parallel window size, model pairs, baseline configuration, and KV-cache handling procedure. The effect of the similarity threshold is analyzed in Appendix~\ref{sec:speedvsaccuracy}, and task-level SpecBench results and adaptive parallel-depth behavior are reported in Appendix~\ref{sec:specbench_results} and Appendix~\ref{sec:parallel_depth}, respectively. The assumptions underlying Theorem~\ref{thm:speculative_bound} are stated explicitly, and a complete proof is provided in Appendix~\ref{sec:proof}. We will provide the implementation and experimental configuration as anonymized supplementary material to facilitate reproduction of the reported results.



\bibliographystyle{iclr2027_conference}
\bibliography{references}

\clearpage
\appendix
\section{Appendix}
\subsection{Proof of Theorem~\ref{thm:speculative_bound}}
\label{sec:proof}

\speculativeBound*

\begin{proof}
Let
\begin{equation}
\delta_r:=\left\|\bar{h}_{m_r}-H_r^*\right\|_2
\end{equation}
denote the deviation between the hidden state produced by ALD and the corresponding hidden state obtained by exact sequential execution at the $r$-th committed depth. By construction, both executions start from the same initial hidden state, so
\begin{equation}
\bar{h}_{m_0}=h_0=H_0^*,
\end{equation}
which gives
\begin{equation}
\delta_0=0.
\end{equation}

Consider an arbitrary committed block $r\in\{0,\ldots,T-1\}$. By definition,
\begin{equation}
F_r=f_{m_{r+1}-1}\circ\cdots\circ f_{m_r}
\end{equation}
maps the hidden state at depth $m_r$ to the hidden state at depth $m_{r+1}$ under exact sequential execution. Hence,
\begin{equation}
H_{r+1}^*=F_r(H_r^*).
\end{equation}
The local approximation error introduced by the corresponding ALD advancement is
\begin{equation}
\epsilon_r=\left\|\bar{h}_{m_{r+1}}-F_r(\bar{h}_{m_r})\right\|_2.
\end{equation}

We can therefore write
\begin{align}
\delta_{r+1}
&=\left\|\bar{h}_{m_{r+1}}-H_{r+1}^*\right\|_2\\
&=\left\|\bar{h}_{m_{r+1}}-F_r(H_r^*)\right\|_2.
\end{align}
Adding and subtracting $F_r(\bar{h}_{m_r})$ inside the norm and applying the triangle inequality yields
\begin{align}
\delta_{r+1}
&\leq\left\|\bar{h}_{m_{r+1}}-F_r(\bar{h}_{m_r})\right\|_2+\left\|F_r(\bar{h}_{m_r})-F_r(H_r^*)\right\|_2\\
&=\epsilon_r+\left\|F_r(\bar{h}_{m_r})-F_r(H_r^*)\right\|_2.
\end{align}

It remains to bound the second term. By assumption, each layer mapping $f_k$ is $L_k$-Lipschitz. Therefore, for any two hidden states $a$ and $b$,
\begin{equation}
\left\|f_k(a)-f_k(b)\right\|_2\leq L_k\left\|a-b\right\|_2.
\end{equation}
Since $F_r$ is the composition of the layers $f_{m_r},\ldots,f_{m_{r+1}-1}$, a Lipschitz constant for $F_r$ is given by the product of the individual layer-wise constants:
\begin{equation}
\Lambda_r=\prod_{k=m_r}^{m_{r+1}-1}L_k.
\end{equation}
Consequently,
\begin{equation}
\left\|F_r(\bar{h}_{m_r})-F_r(H_r^*)\right\|_2\leq\Lambda_r\left\|\bar{h}_{m_r}-H_r^*\right\|_2=\Lambda_r\delta_r.
\end{equation}
Combining the above inequalities gives the recursive bound
\begin{equation}
\delta_{r+1}\leq\Lambda_r\delta_r+\epsilon_r.
\end{equation}

We now unroll this recursion. Since $\delta_0=0$,
\begin{align}
\delta_1&\leq\epsilon_0,\\
\delta_2&\leq\Lambda_1\delta_1+\epsilon_1\leq\Lambda_1\epsilon_0+\epsilon_1,\\
\delta_3&\leq\Lambda_2\delta_2+\epsilon_2\leq\Lambda_2\Lambda_1\epsilon_0+\Lambda_2\epsilon_1+\epsilon_2.
\end{align}
Continuing recursively, after $T$ committed blocks, we obtain
\begin{equation}
\delta_T\leq\sum_{r=0}^{T-1}\left(\prod_{q=r+1}^{T-1}\Lambda_q\right)\epsilon_r.
\end{equation}
The product is interpreted as $1$ when $r=T-1$, corresponding to the fact that the error introduced in the final committed block is not propagated through any subsequent block.

Finally, since $m_T=L$, we have
\begin{equation}
\delta_T=\left\|\bar{h}_L-H_T^*\right\|_2.
\end{equation}
Therefore,
\begin{equation}
\left\|\bar{h}_L-H_T^*\right\|_2\leq\sum_{r=0}^{T-1}\left(\prod_{q=r+1}^{T-1}\Lambda_q\right)\epsilon_r,
\end{equation}
which proves the result.
\end{proof}

\subsection{Task-Level Results on SpecBench}
\label{sec:specbench_results}

\begin{table}[t]
\centering
\caption{Acceptance rate and end-to-end decoding speedup on SpecBench. E2E speedup is measured relative to standard autoregressive decoding with the corresponding target model. The draft models are Qwen-4B and LLaMA-3B for the Qwen-30B and LLaMA-70B target models, respectively.}
\label{tab:task-level-results}
\small
\setlength{\tabcolsep}{6pt}
\begin{tabular}{llcccc}
\toprule
\multirow{2}{*}{Task} & \multirow{2}{*}{Method} & \multicolumn{2}{c}{Qwen-4B $\rightarrow$ 30B} & \multicolumn{2}{c}{LLaMA-3B $\rightarrow$ 70B} \\
\cmidrule(lr){3-4} \cmidrule(lr){5-6}
& & Accept. & E2E Spd. & Accept. & E2E Spd. \\
\midrule
\multirow{4}{*}{MT}
& SD        & \textbf{0.38} & 2.15$\times$ & \textbf{0.47} & 1.98$\times$ \\
& EasySpec  & 0.31 & 2.52$\times$ & 0.39 & 2.18$\times$ \\
& LayerSkip & 0.26 & 2.31$\times$ & 0.36 & 2.05$\times$ \\
& ALD       & 0.35 & \textbf{2.75$\times$} & 0.42 & \textbf{2.25$\times$} \\
\midrule
\multirow{4}{*}{TL}
& SD        & \textbf{0.41} & 2.34$\times$ & \textbf{0.41} & 1.84$\times$ \\
& EasySpec  & 0.34 & 2.67$\times$ & 0.34 & 1.94$\times$ \\
& LayerSkip & 0.30 & 2.45$\times$ & 0.28 & 1.98$\times$ \\
& ALD       & 0.36 & \textbf{2.91$\times$} & 0.40 & \textbf{2.01$\times$} \\
\midrule
\multirow{4}{*}{SUM}
& SD        & \textbf{0.34} & 1.97$\times$ & \textbf{0.42} & 1.76$\times$ \\
& EasySpec  & 0.29 & 2.13$\times$ & 0.38 & 1.91$\times$ \\
& LayerSkip & 0.24 & 2.08$\times$ & 0.31 & 1.83$\times$ \\
& ALD       & 0.33 & \textbf{2.48$\times$} & 0.39 & \textbf{2.28$\times$} \\
\midrule
\multirow{4}{*}{QA}
& SD        & \textbf{0.32} & 1.76$\times$ & \textbf{0.36} & 1.68$\times$ \\
& EasySpec  & 0.31 & 1.85$\times$ & 0.27 & 1.79$\times$ \\
& LayerSkip & 0.19 & 1.86$\times$ & 0.24 & 1.71$\times$ \\
& ALD       & 0.30 & \textbf{2.09$\times$} & 0.32 & \textbf{2.11$\times$} \\
\midrule
\multirow{4}{*}{MR}
& SD        & \textbf{0.53} & 3.14$\times$ & \textbf{0.68} & 2.67$\times$ \\
& EasySpec  & 0.45 & 3.22$\times$ & 0.61 & 2.74$\times$ \\
& LayerSkip & 0.35 & 3.20$\times$ & 0.52 & 2.70$\times$ \\
& ALD       & 0.48 & \textbf{3.42$\times$} & 0.67 & \textbf{2.97$\times$} \\
\midrule
\multirow{4}{*}{RAG}
& SD        & \textbf{0.35} & 2.09$\times$ & \textbf{0.59} & 2.17$\times$ \\
& EasySpec  & 0.30 & 2.30$\times$ & 0.54 & 2.29$\times$ \\
& LayerSkip & 0.23 & 2.16$\times$ & 0.48 & 2.26$\times$ \\
& ALD       & 0.34 & \textbf{2.51$\times$} & 0.56 & \textbf{2.49$\times$} \\
\bottomrule
\end{tabular}
\end{table}

Table~\ref{tab:task-level-results} evaluates ALD across the heterogeneous generation tasks included in SpecBench. Across both model pairs, ALD improves end-to-end decoding speedup over standard SD and the compared draft-acceleration baselines while retaining acceptance rates closer to standard SD than the more aggressive approximations.

The advantage is particularly evident in settings where preserving draft quality is critical to end-to-end efficiency. For example, on mathematical reasoning with LLaMA-3B $\rightarrow$ LLaMA-70B, standard SD obtains an acceptance rate of $0.68$ and a $2.67\times$ speedup. EasySpec increases speedup to $2.74\times$ while reducing acceptance to $0.61$, whereas ALD retains an acceptance rate of $0.67$ and reaches $2.97\times$ speedup. Similarly, on summarization with Qwen-4B $\rightarrow$ Qwen-30B, ALD achieves $2.48\times$ speedup with an acceptance rate of $0.33$, compared with $2.13\times$ and $0.29$ for EasySpec.

These results illustrate why drafting latency alone is insufficient to characterize the efficiency of speculative decoding. A more aggressive approximation can reduce the cost of producing speculative tokens while simultaneously increasing rejection frequency, thereby reducing the amount of useful work obtained from each target-model verification. ALD instead uses the online similarity criterion to regulate the extent of proxy execution, retaining greater draft--target agreement while exposing substantial layer-level parallelism.
The task-level variation motivates adaptive drafting. Different generation tasks induce different representation dynamics and draft--target agreement, so a uniform approximation strategy need not provide the same latency--acceptance trade-off across all inputs. ALD responds to these variations via its online hidden-state similarity criterion, without requiring task-specific retraining or architectural modifications.

\subsection{Effect of the Similarity Threshold}
\label{sec:speedvsaccuracy}

\begin{figure}[t]
\centering
\includegraphics[width=0.5\linewidth]{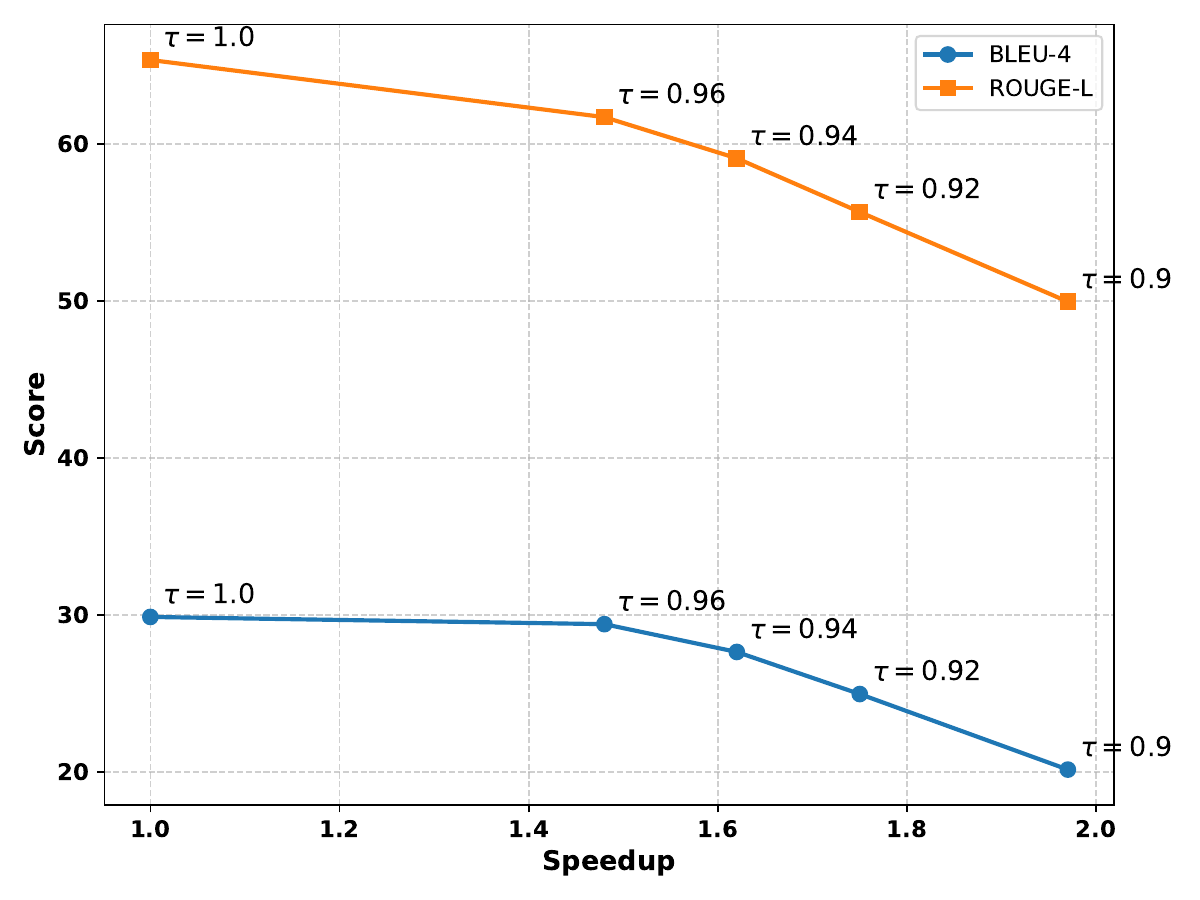}
\caption{Effect of the similarity threshold $\tau$ on drafting speedup and draft-model quality, measured using BLEU-4 and ROUGE-L. Lower $\tau$ permits more aggressive layer-parallel proxy execution and increases speedup, while higher $\tau$ produces more conservative execution and better preserves draft quality. The default setting $\tau=0.95$ provides a favorable operating point between these objectives.}
\label{fig:speed_vs_quality}
\end{figure}

Figure~\ref{fig:speed_vs_quality} studies the effect of the similarity threshold $\tau$, which directly controls how conservatively ALD advances through a layer-parallel window. Increasing $\tau$ causes ALD to refresh the committed representation more frequently, reducing effective parallel advancement but keeping the execution closer to standard sequential draft inference. Conversely, decreasing $\tau$ allows ALD to advance farther using proxy representations, increasing parallelism at the cost of greater approximation.

The resulting behavior produces a speed--quality trade-off. At high values of $\tau$, ALD behaves conservatively: fewer proxy transitions are retained, the effective sequential depth increases, and the resulting draft representation remains closer to that of the original draft model. As $\tau$ decreases, more candidate transitions are retained within each parallel window, reducing the number of sequential drafting stages and increasing drafting speedup. A reduction in draft-model quality accompanies this additional acceleration, as increasingly aggressive proxy execution deviates further from standard sequential propagation.

We use $\tau=0.95$ as the default operating point throughout the main experiments. Importantly, this value is shared across models and datasets rather than tuned separately for each benchmark. At this operating point, ALD provides substantial drafting acceleration while maintaining draft quality and verifier acceptance close to standard SD. The use of a common threshold also reduces tuning overhead and evaluates whether a single operating point can transfer across the considered settings.

The behavior at lower thresholds further motivates the adaptive similarity check. If the threshold is excessively permissive, ALD retains proxy transitions even when the corresponding layer transformation substantially changes the representation. Such aggressive advancement can increase approximation error and degrade the resulting draft distribution. The similarity threshold, therefore, acts as a control parameter that regulates how much layer-level parallelism is exposed before the committed representation is refreshed.

The cosine score should be interpreted as an inexpensive online control signal rather than a direct estimator of the Euclidean approximation error appearing in Theorem~\ref{thm:speculative_bound}. The theorem characterizes how local errors arising from ALD propagate through subsequent layer blocks, while Figure~\ref{fig:speed_vs_quality} empirically demonstrates how varying the similarity criterion governs the practical trade-off between approximation and drafting latency.

\subsection{Adaptive Parallel-Depth Behavior}
\label{sec:parallel_depth}

\begin{table}[t]
\centering
\caption{Average reduction in effective sequential layer depth achieved by ALD across datasets. Values are reported as parallelized layer advances relative to the total number of draft-model layers. Larger values indicate that ALD advances farther through the model using layer-parallel proxy execution before refreshing the committed representation.}
\label{tab:layer_skip}
\small
\setlength{\tabcolsep}{6pt}
\renewcommand{\arraystretch}{1.15}
\begin{tabular}{lccccc}
\toprule
\textbf{Model} & \textbf{WMT} & \textbf{MMLU} & \textbf{MGSM} & \textbf{IF-Eval} & \textbf{Math} \\
\midrule
Qwen-4B  & 15.2 / 36 & 18.5 / 36 & 14.8 / 36 & 15.4 / 36 & 16.3 / 36 \\
LLaMA-3B & 12.9 / 28 & 13.6 / 28 & 10.8 / 28 & 13.7 / 28 & 12.0 / 28 \\
\bottomrule
\end{tabular}
\end{table}

To better understand the adaptive behavior of ALD, Table~\ref{tab:layer_skip} reports the average reduction in effective sequential depth across datasets. For Qwen-4B, which contains $36$ transformer layers, ALD achieves average parallel advances corresponding to $15.2$ layers on WMT, $18.5$ on MMLU, $14.8$ on MGSM, $15.4$ on IF-Eval, and $16.3$ on Math. For LLaMA-3B, which contains $28$ layers, the corresponding values are $12.9$, $13.6$, $10.8$, $13.7$, and $12.0$ layers.

These results show that ALD substantially reduces the effective sequential depth of the draft model across different tasks. The amount of parallel advancement is input- and task-dependent: representationally stable regions allow ALD to advance farther within a parallel window, whereas larger hidden-state changes trigger earlier updates of the committed representation. This variation is the behavior that the adaptive similarity criterion targets.

For Qwen-4B, the largest average effective-depth reduction occurs on MMLU, where ALD advances across an equivalent of $18.5$ out of $36$ layers through layer-parallel proxy execution. At the same time, MGSM exhibits a smaller reduction of $14.8$ layers. A similar variation is observed for LLaMA-3B, where the average reduction ranges from $10.8$ layers on MGSM to $13.7$ on IF-Eval. These differences indicate that ALD does not impose a uniform execution pattern across tasks.

Importantly, the values in Table~\ref{tab:layer_skip} represent reductions in \emph{sequential dependency depth}, rather than transformer layers physically removed from the model. The corresponding layer transformations are evaluated through the proxy execution described in Section~\ref{sec:drafting}; acceleration arises from evaluating multiple transformations concurrently and reducing the number of sequential stages required to obtain the final draft representation.

\subsection{Relation to Training-Based Draft Optimization}
\label{sec:training_based}

A complementary line of work designs draft models explicitly optimized for speculative decoding through additional training, including EAGLE-style approaches such as EAGLE-3. These methods reduce drafting cost by learning lightweight predictors or specialized draft architectures that produce speculative representations or tokens more efficiently than a conventional autoregressive drafter.

Training-based draft optimization operates under different deployment assumptions from ALD. Such approaches require an additional optimization procedure and access to suitable training data, model features, or distillation pipelines. In contrast, ALD operates directly on an existing multi-layer draft model without modifying its parameters or architecture. Its objective is therefore not to learn a new drafter, but to change how the layers of an existing drafter are executed at inference time.

The two directions are complementary. A trained lightweight drafter reduces the intrinsic cost of speculative proposal generation, whereas ALD targets sequential dependencies within an existing multi-layer drafter. The potential benefit of ALD consequently depends on the depth and computational structure of the draft model. When the drafter is already extremely shallow, there is less layer-level parallelism available to exploit. Conversely, for off-the-shelf multi-layer draft models, ALD provides a training-free mechanism for reducing sequential drafting latency.

We focus our main comparisons on methods that operate in the training-free or layer-reduction setting targeted by ALD. Investigating whether adaptive layer-parallel execution can be combined with learned speculative drafters is an interesting direction for future work.

\subsection{Additional Implementation and Reproducibility Details}
\label{sec:implementation_details}

All experiments use a speculative draft length of $6$, similarity threshold $\tau=0.95$, and maximum parallel window $K=10$, unless otherwise stated. ALD uses the same pretrained draft and target models as standard speculative decoding and requires no parameter updates, auxiliary prediction heads, or task-specific fine-tuning. The similarity check is computed online from the hidden representations produced during draft execution.

For each draft token, ALD begins from the most recently committed hidden representation. The first layer transformation from this state is evaluated exactly, while the remaining transformations within the current window are evaluated concurrently using the committed representation as their shared proxy input. The cosine-similarity criterion determines how far the resulting proxy execution is advanced before the committed representation is refreshed. This procedure is repeated until the final draft model depth is reached.

Following target-model verification, approximate KV states produced during proxy execution are discarded and reconstructed using the bonus-calibration procedure described in Section~\ref{sec:kv_alignment}. This prevents approximate KV states from persisting across speculative decoding iterations. All reported end-to-end measurements use the same verification and KV-cache handling procedure for the corresponding method throughout an experiment.

Unless otherwise stated, hyperparameters for baseline methods follow their official implementations. We use identical model pairs, speculative draft lengths, datasets, and hardware when comparing methods to minimize differences unrelated to the drafting strategy.\\

\subsection{Sensitivity to the Maximum Parallel Window $K$}
\label{app:k_sensitivity}

The maximum parallel window $K$ determines the number of future
layer transformations that ALD may evaluate concurrently from the
current committed representation. In all main experiments, we use
$K=10$. We additionally study the sensitivity of ALD to this choice
by varying $K$ while keeping the remaining inference configuration
fixed.

Importantly, $K$ specifies an \emph{upper bound} on the available
layer parallelism rather than the number of proxy transitions that
are necessarily committed. Within each parallel window, the
similarity criterion continues to determine how far ALD advances.
If the hidden-state similarity falls below the threshold $\tau$,
the remaining candidates are discarded and execution resumes from
the last accepted representation. Consequently, increasing $K$
exposes additional candidate parallelism without forcing ALD to use
increasingly distant proxy states.

\begin{table}[t]
    \centering
    \begin{tabular}{c|ccc}
        \toprule
        $K$ &
        Acceptance Rate &
        E2E Speedup &
        Draft Time (s/100 tok.) \\
        \midrule
        5     & 0.42 & $2.81\times$ & 20.4 \\
        10    & 0.38 & $2.95\times$ & 18.7 \\
        15    & 0.36 & $2.99\times$ & 18.2 \\
        20    & 0.35 & $3.01\times$ & 18.0 \\
        $L-1$ & 0.34 & $3.03\times$ & 17.7 \\
        \bottomrule
    \end{tabular}
    \caption{
    Sensitivity to the maximum parallel window $K$.
    Increasing $K$ exposes more candidate layer transformations to
    parallel execution. The similarity criterion limits how much of
    this additional parallelism is actually committed, leading to
    diminishing speedup gains beyond $K=10$.
    }
    \label{tab:k_sensitivity}
\end{table}

Increasing $K$ from 5 to 10 reduces draft time from 20.4 to
18.7 seconds per 100 tokens and improves end-to-end speedup from
$2.81\times$ to $2.95\times$. Beyond this point, the gains
saturate: using the entire remaining model depth ($K=L-1$) improves
speedup only marginally further to $3.03\times$. At the same time,
acceptance decreases gradually from 0.42 at $K=5$ to 0.34 at
$K=L-1$.

This behavior is consistent with the adaptive design of ALD. A
parallel window rarely advances through substantially more than
approximately ten layers before representational divergence causes
the similarity criterion to refresh the committed state. Additional
candidates at larger $K$ are therefore increasingly likely to be
discarded, yielding limited latency reduction while introducing
additional parallel computation.

These results motivate $K=10$ as the default configuration used
throughout the main experiments. It captures most of the available
layer-level parallelism while avoiding unnecessary candidate
computation, and we use the same value across tasks without
per-dataset tuning.

\subsection{System-Level Profiling and Overhead Analysis}
\label{app:system_profiling}

ALD trades additional parallel computation within the draft model
for reduced sequential execution depth. We therefore complement the
end-to-end speedups in Section~5 with a detailed characterization of
GPU utilization, memory overhead, synchronization cost, and the
runtime of individual ALD operations. For this profiling experiment, we use the Qwen-4B drafter on an
NVIDIA Tesla T4 GPU and representative prompts spanning translation,
summarization, question answering, and reasoning. The T4 provides a
substantially more resource-constrained single-GPU environment than
the multi-GPU setup used for the main experiments, allowing us to
evaluate ALD when computational resources are limited.

\subsubsection{GPU Utilization}
\label{app:gpu_utilization}

We first compare GPU utilization during standard sequential drafting
and ALD.

\begin{table}[t]
    \centering
    \begin{tabular}{c|ccc}
        \toprule
        Batch Size &
        Vanilla Drafting &
        ALD &
        Improvement \\
        \midrule
        1 & 43.5\% & \textbf{57.4\%} & +13.9 pts \\
        8 & 87.9\% & \textbf{98.4\%} & +10.5 pts \\
        \bottomrule
    \end{tabular}
    \caption{
    GPU utilization during draft generation. ALD increases
    utilization even at batch size one, indicating that its benefit
    does not depend exclusively on request-level batching.
    }
    \label{tab:gpu_utilization}
\end{table}

At batch size one, standard sequential drafting utilizes 43.5\% of
the GPU, whereas ALD increases utilization to 57.4\%, an absolute
improvement of 13.9 percentage points. The improvement therefore
does not rely on batching multiple independent requests: ALD exposes
parallelism within the transformer-layer execution of a single
request. At batch size eight, utilization further increases from
87.9\% to 98.4\%.

These measurements support the system-level motivation of ALD.
Sequential layer execution can leave substantial compute capacity
unused during drafting, and layer-parallel execution converts part
of this idle capacity into useful computation.

\subsubsection{Memory Overhead}
\label{app:memory_overhead}

ALD does not duplicate model parameters or introduce additional
transformer layers. Its additional memory usage consists of
transient activations and temporary buffers required by parallel
proxy evaluations and the similarity check.

We measure the peak additional GPU memory required for one complete
ALD drafting round, including proxy construction, $K$ parallel
candidate evaluations, similarity computation, correction, and
KV-cache reconstruction.

\begin{table}[t]
    \centering
    \begin{tabular}{c|cc}
        \toprule
        Batch Size &
        Peak Additional Memory &
        \% of Drafter Weights \\
        \midrule
        1 & 2.7 MB  & 0.03\% \\
        8 & 22.0 MB & 0.27\% \\
        \bottomrule
    \end{tabular}
    \caption{
    Peak additional GPU memory introduced by ALD. The percentage
    is measured relative to the approximately 8\,GB drafter weight
    footprint.
    }
    \label{tab:memory_overhead}
\end{table}

At batch size one, ALD requires only 2.7\,MB of additional peak
memory, corresponding to approximately 0.03\% of the drafter's
weight footprint. At batch size eight, this increases to 22.0\,MB,
or approximately 0.27\%. The overhead therefore remains below 1\%
of the drafter weight footprint in both evaluated settings.

The additional memory scales with the number of concurrently
processed activations and is released after each drafting round.
Moreover, ALD modifies only draft execution; target-model
verification follows the standard speculative-decoding procedure
and incurs no ALD-specific memory overhead.

\subsubsection{Synchronization and Runtime Breakdown}
\label{app:runtime_breakdown}

ALD introduces one synchronization point per parallel drafting
iteration. After the $K$ proxy layer evaluations complete,
synchronization is required before computing their similarities and
selecting the deepest accepted candidate. Since the proxy
evaluations correspond to transformer layers with identical
architecture and similar computational workloads, they exhibit
limited load imbalance and generally complete at similar times.

We profile the principal components of one ALD iteration in
Table~\ref{tab:runtime_breakdown}.

\begin{table}[t]
    \centering
    \begin{tabular}{l|cc}
        \toprule
        Stage & Batch 1 & Batch 8 \\
        \midrule
        Proxy construction
            & 0.33 ms & 0.31 ms \\
        Proxy layer evaluation ($K=10$)
            & 11.8 ms & 41.9 ms \\
        Similarity check
            & 2.0 ms & 2.3 ms \\
        Vanilla single-layer forward
            & 5.8 ms & 15.6 ms \\
        \bottomrule
    \end{tabular}
    \caption{
    Runtime breakdown of ALD operations. Proxy layer evaluation
    evaluates up to $K=10$ candidate layer transformations
    concurrently.
    }
    \label{tab:runtime_breakdown}
\end{table}

A complete ALD iteration costs approximately $2.4$--$2.7\times$
the latency of a single sequential layer transition while evaluating
up to $K=10$ candidate transformations concurrently. The resulting
break-even point is therefore reached when an iteration advances
through approximately 2--3 layer transitions. Beyond this point,
replacing several sequential layer evaluations with one parallel
ALD round reduces sequential latency.

The adaptive-depth measurements in Appendix~A.4 show that ALD
advances substantially farther than this break-even point on average
across the evaluated tasks, explaining why the additional
bookkeeping and synchronization costs are amortized in practice.

Finally, the KV-cache reconstruction described in Section~4.3 is
included in all reported end-to-end measurements. This operation is
required to realign the draft cache after approximate execution and
has latency comparable to generating a single token in our
implementation. Thus, the speedups reported in Section~5 represent
\emph{net end-to-end improvements}, including proxy construction,
similarity evaluation, synchronization, correction, and KV-cache
reconstruction.

\subsection{Robustness Across Repeated Runs and Random Seeds}
\label{app:robustness}

To distinguish runtime variability from sequence-dependent
variation, we perform two complementary five-repetition studies:
(i) repeated execution using an identical random seed and
(ii) independent executions using different random seeds.

\subsubsection{Repeated Execution with the Same Seed}

\begin{table}[t]
    \centering
    \begin{tabular}{l|cc}
        \toprule
        Dataset &
        Acceptance Rate &
        E2E Speedup \\
        \midrule
        WMT  & $0.38 \pm 0.0000$ & $2.95 \pm 0.0057\times$ \\
        MMLU & $0.36 \pm 0.0000$ & $2.77 \pm 0.0037\times$ \\
        MGSM & $0.35 \pm 0.0000$ & $2.61 \pm 0.0077\times$ \\
        Math & $0.39 \pm 0.0000$ & $3.75 \pm 0.0087\times$ \\
        \bottomrule
    \end{tabular}
    \caption{
    Repeated ALD measurements with a fixed random seed.
    Values report mean $\pm$ standard deviation over five runs.
    }
    \label{tab:same_seed}
\end{table}

At a fixed seed, the acceptance decisions are reproducible,
resulting in zero observed standard deviation in acceptance rate.
End-to-end latency exhibits only small system-level variation: the
standard deviation in measured speedup remains below $0.009\times$
across all four benchmarks. This residual variation reflects
ordinary runtime effects such as GPU kernel scheduling rather than
instability in the ALD decision procedure.

\subsubsection{Variation Across Random Seeds}

We next repeat the experiment with independently sampled random
seeds.

\begin{table}[t]
    \centering
    \begin{tabular}{l|cc}
        \toprule
        Dataset &
        Acceptance Rate &
        E2E Speedup \\
        \midrule
        WMT  & $0.39 \pm 0.1062$ & $2.95 \pm 0.2165\times$ \\
        MMLU & $0.36 \pm 0.0663$ & $2.76 \pm 0.1559\times$ \\
        MGSM & $0.35 \pm 0.0679$ & $2.60 \pm 0.0816\times$ \\
        Math & $0.38 \pm 0.0951$ & $3.77 \pm 0.2538\times$ \\
        \bottomrule
    \end{tabular}
    \caption{
    Robustness across random seeds. Values report mean $\pm$
    standard deviation over five independent seeds.
    }
    \label{tab:different_seeds}
\end{table}

As expected, variability is larger across seeds because different
sampled sequences induce different hidden-state trajectories and
therefore different opportunities for adaptive layer-parallel
execution. Nevertheless, the mean acceptance rates and end-to-end
speedups remain closely aligned with those reported in the main
experiments.

Together, these results separate two sources of variation. Repeated
execution of the same decoding trajectory is highly reproducible,
whereas variation across seeds reflects genuine differences in the
generated sequences rather than instability of the ALD execution
mechanism.

\subsection{Generalization to the Gemma Model Family}
\label{app:gemma}

The main experiments evaluate ALD using Qwen and LLaMA model pairs.
To examine whether the observed layer-parallel behavior extends
beyond these two model families, we additionally evaluate ALD using
Gemma-3-1B as the draft model and Gemma-3-27B as the target model.

\begin{table}[t]
    \centering
    \begin{tabular}{ll|ccc}
        \toprule
        Dataset &
        Method &
        Acceptance &
        E2E Speedup &
        Draft Time (s/100 tok.) \\
        \midrule
        \multirow{2}{*}{MMLU}
          & SD  & 0.45 & $1.79\times$ & 8.5 ($1.00\times$) \\
          & ALD & 0.41 & $\mathbf{2.05\times}$ &
            \textbf{5.1 ($1.67\times$)} \\
        \midrule
        \multirow{2}{*}{Math}
          & SD  & 0.52 & $1.93\times$ & 8.2 ($1.00\times$) \\
          & ALD & 0.50 & $\mathbf{2.15\times}$ &
            \textbf{4.7 ($1.75\times$)} \\
        \bottomrule
    \end{tabular}
    \caption{
    Generalization to the Gemma model family using
    Gemma-3-1B $\rightarrow$ Gemma-3-27B. Draft-time values in
    parentheses denote acceleration relative to standard SD
    drafting.
    }
    \label{tab:gemma}
\end{table}

On MMLU, ALD reduces draft time from 8.5 to 5.1 seconds per
100 tokens, corresponding to a $1.67\times$ drafting acceleration.
This improves end-to-end speedup from $1.79\times$ under standard
SD to $2.05\times$, while acceptance changes from 0.45 to 0.41.

A similar pattern appears on Math. Draft time decreases from 8.2 to
4.7 seconds per 100 tokens ($1.75\times$), and end-to-end speedup
increases from $1.93\times$ to $2.15\times$, while acceptance
remains close to standard SD (0.50 versus 0.52).

These results provide additional evidence that the layer-level
parallelism exploited by ALD is not specific to the Qwen and LLaMA
architectures used in the main experiments. Nevertheless, the
magnitude of the benefit depends on the depth and representation
dynamics of the particular draft model, as discussed in
Appendix~A.5.

\subsection{Similarity Metric Ablation}
\label{app:similarity_metric}

ALD uses cosine similarity as an inexpensive online signal for
deciding whether proxy layer transitions remain sufficiently stable
to be retained. We additionally evaluate mean squared error (MSE)
between hidden representations as an alternative similarity
criterion.

Both criteria provide a signal of representational deviation, but
they differ in an important respect. MSE depends directly on the
absolute magnitude of the hidden activations and was empirically
more sensitive to changes in representation scale across decoding
positions and model configurations. Consequently, a fixed MSE
threshold was less stable across the evaluated settings.

Cosine similarity instead measures directional agreement,
\begin{equation}
s(\mathbf{h},\tilde{\mathbf{h}})
=
\frac{
    \langle \mathbf{h},\tilde{\mathbf{h}} \rangle
}{
    \|\mathbf{h}\|_2
    \|\tilde{\mathbf{h}}\|_2
},
\end{equation}
and is invariant to the norms of the two representations. Since ALD
seeks an inexpensive indicator of whether hidden representations
continue to evolve in a similar direction under proxy execution,
this normalization makes the criterion less sensitive to changes in
activation magnitude. In our experiments, MSE produced qualitatively
comparable behavior but required greater sensitivity in threshold
selection, whereas cosine similarity provided a more stable
operating criterion across decoding steps and model scales.

We emphasize that neither criterion constitutes a correctness
certificate for the approximate hidden state. In particular, two
representations can have high cosine similarity while differing in
fine-grained components relevant to subsequent computation. We
therefore interpret cosine similarity as an \emph{online control
signal} for adaptive execution rather than as a direct estimator of
the Euclidean approximation error $\epsilon_r$ in Theorem~1. When
the similarity falls below the threshold, ALD refreshes the
committed representation and begins a new parallel window.

This interpretation is consistent with Appendix~A.3, where varying
$\tau$ empirically controls the trade-off between parallel
advancement and draft quality, and with Theorem~1, which separately
characterizes how local hidden-state approximation errors propagate
through subsequent layer blocks.


\end{document}